\documentclass[final]{l4dc2026}

\title[RoVer-CoRe]{Robust Verification of Controllers under State Uncertainty via Hamilton-Jacobi Reachability Analysis}
\usepackage{times}
\usepackage{sidecap}
\usepackage{mathtools}
\usepackage{amsmath}
\usepackage{wrapfig}
\DeclareMathOperator{\sgn}{sgn}

\author{%
 \Name{Albert Lin} \Email{albertkl@stanford.edu}\\
 \addr Stanford University, CA, US
 \AND
 \Name{Alessandro Pinto} \Email{alessandro.pinto@jpl.nasa.gov}\\
 \addr NASA Jet Propulsion Laboratory, CA, US
 \AND
 \Name{Somil Bansal} \Email{somil@stanford.edu}\\
 \addr Stanford University, CA, US%
}

\begin{document}

\maketitle

\begin{abstract}%
  As perception-based controllers for autonomous systems become increasingly popular in the real world, it is important that we can formally verify their safety and performance despite \textit{perceptual uncertainty}.
  Unfortunately, the verification of such systems remains challenging, largely due to the complexity of the controllers, which are often nonlinear, nonconvex, learning-based, and/or black-box.
  Prior works propose verification algorithms that are based on approximate reachability methods, but they often restrict the class of controllers and systems that can be handled or result in overly conservative analyses.
  Hamilton-Jacobi (HJ) reachability analysis is a popular formal verification tool for general nonlinear systems that can compute optimal reachable sets under worst-case system uncertainties; however, its application to perception-based systems is currently underexplored.
  In this work, we propose RoVer-CoRe, a framework for the \underline{\textbf{Ro}}bust \underline{\textbf{Ver}}ification of \underline{\textbf{Co}}ntrollers via HJ \underline{\textbf{Re}}achability.
  To the best of our knowledge, RoVer-CoRe is the first HJ reachability-based framework for the verification of perception-based systems under perceptual uncertainty.
  Our key insight is to concatenate the system controller, observation function, and the state estimation modules to obtain an equivalent closed-loop system that is readily compatible with existing reachability frameworks.
  Within RoVer-CoRe, we propose novel methods for formal safety verification and robust controller design.
  We demonstrate the efficacy of the framework in case studies involving aircraft taxiing and NN-based rover navigation.
  Code is available at the link in the footnote\footnote{\label{note:code}\url{https://github.com/albertklin/rover-core}}.%
\end{abstract}


\section{Introduction}\label{sec:introduction}

As perception-based controllers for autonomous systems become increasingly popular in the real world, it is important that we can formally verify their safety and performance despite \textit{perceptual uncertainty}.
Perceptual uncertainty can result from inherent noise in the robot's sensors, inevitable learning errors in learning-based modules, as well as environmental factors, e.g., when a tunnel blocks the reception of a GPS signal.
In this work, we are interested in computing formal guarantees for such systems despite worst-case perceptual errors.
For example, we would like to verify that a rover will not enter safety-critical keep-out zones even if it encounters adversarial errors in a vision-based state estimation module.
Unfortunately, verifying perception-based systems remains challenging, largely due to the complexity of the controllers, which are often nonlinear, nonconvex, learning-based, and/or black-box \citep{edwards2024generalframeworkverificationcontrol}.
These qualities make the formal analysis of controllers a difficult problem on its own, even without perceptual uncertainty.

\begin{figure}[!ht]
\centering
\includegraphics[width=0.55\textwidth]{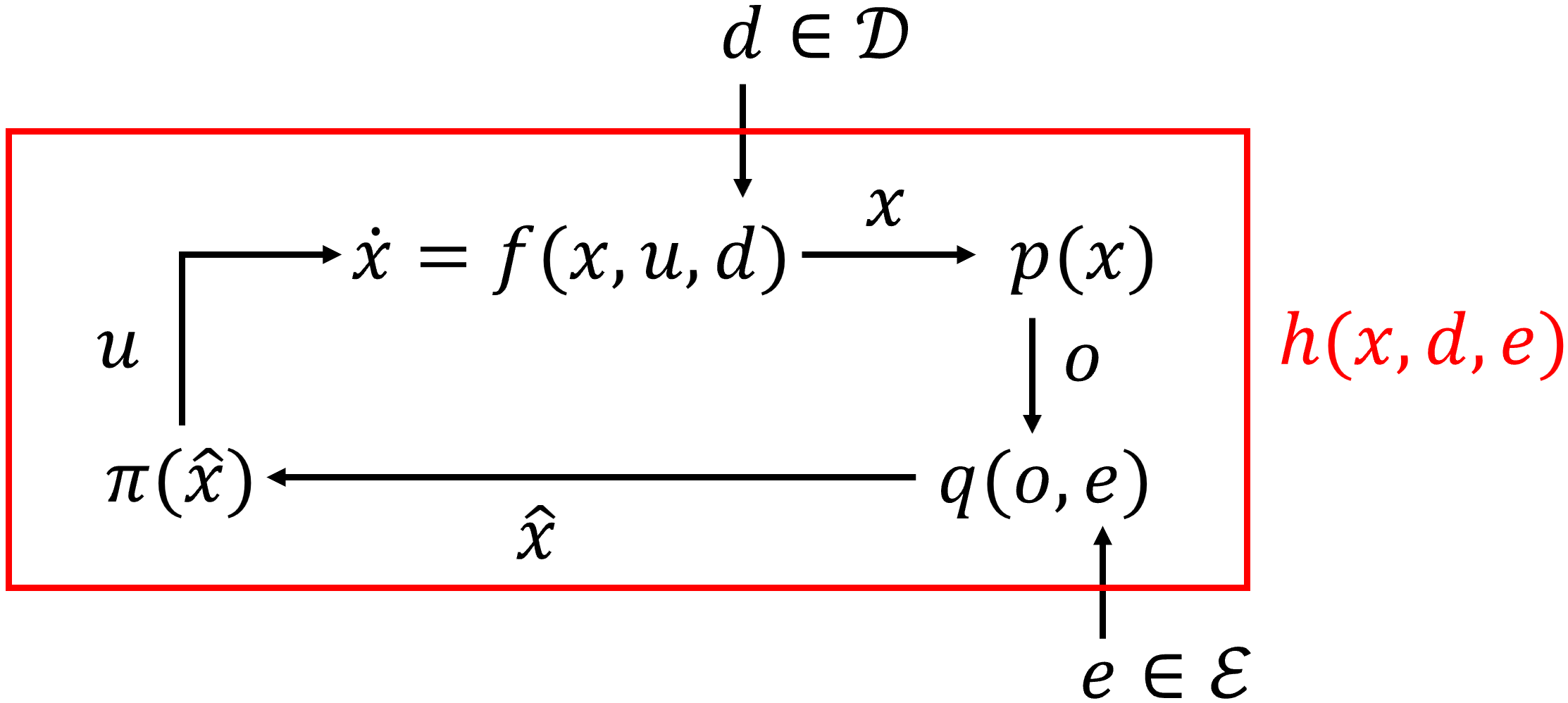}
\caption{(Black) A perception-based system under perceptual uncertainty $e$, which result in noisy state estimates $\hat{x}$. {\color{red}(Red)} The closed-loop system abstraction used for safety analysis.}
\label{fig:system}
\end{figure}

Since many perception-based controllers are implemented with neural networks (NN) due to their ability to process general inputs, a substantial body of prior works focuses on verifying such systems by combining NN verification tools with control-theoretic verification methods \citep{10.1007/978-3-031-65112-0_5,trapiello2023verificationneuralnetworkcontrol,wang2024modelfreeverificationneuralnetwork,9683154,9502456,10.1145/3501710.3519511,ROSSI2024109238,9683154}.
The most common approaches leverage approximate reachable set methods to verify the closed-loop system as a whole \citep{9502456,10558853,9992984,DUTTA2018151,schilling,10.1145/3387168.3387244,xiang2018reachability,ROSSI2024109238,10.1609/aaai.v37i12.26783,10.1007/978-3-030-53288-8_1,10.1145/3302504.3311806,10.1007/978-3-030-59152-6_30,9561348,9992719}.
However, such methods are heavily constrained by the conservatism of the chosen propagation tools, which can result in a failure to certify system safety.
We propose to overcome this limitation by using Hamilton-Jacobi (HJ) reachability analysis for the reachability computation \citep{bansal2017hamilton}.
Our key insight is that HJ reachability can provide a less conservative result than existing approaches because it computes the optimal reachable sets for general nonlinear dynamical systems.

Building on this insight, we propose RoVer-CoRe, a framework for the \underline{\textbf{Ro}}bust \underline{\textbf{Ver}}ification of \underline{\textbf{Co}}ntrollers via HJ \underline{\textbf{Re}}achability.
RoVer-CoRe concatenates the system controller, observation function, and the state estimation modules to obtain an equivalent closed-loop system, where the only external input is perceptual uncertainty.
This equivalent closed-loop system is readily handled under existing HJ reachability tools after appropriate adjustments.
Within RoVer-CoRe, we show that the main challenge in applying HJ reachability tools comes from computing the closed-loop system Hamiltonian.
To overcome this challenge, we propose different methods to bound the Hamiltonian depending on the form of the controller, resulting in potentially conservative yet sound analyses.
Finally, we demonstrate the efficacy of RoVer-CoRe for formal safety verification and robust controller design on case studies involving aircraft taxiing and NN-based rover navigation.
To our knowledge, RoVer-CoRe is the first HJ reachability-based framework for verifying perception-based systems.
In summary, our contributions include:
\begin{itemize}\setlength\itemsep{-0.1em}
    \item RoVer-CoRe, an HJ reachability-based framework for verifying perception-based systems,
    \item methods for formal safety verification and robust controller design under uncertainty, and
    \item case studies demonstrating the efficacy of the proposed framework, with code open-sourced\textsuperscript{\ref{note:code}}.
\end{itemize}
\section{Related Works}\label{sec:related_works}

\subsection{Approximate Reachable Set Methods}
Previous works combine NN verification tools with approximate reachable set methods to verify NN-controlled systems under perceptual uncertainty \citep{9683154}.
In such works, NN verification tools bound the closed-loop system dynamics subject to bounded perturbations in the controller's observations.
Then, approximate reachable sets can be computed using the bounded dynamics.

To efficiently propagate sets for nonlinear systems, approximate methods employ various finite set representations, such as hyper-rectangles \citep{9502456,xiang2018reachability,ROSSI2024109238}, polytopes \citep{DUTTA2018151,10.1145/3387168.3387244,10.1609/aaai.v37i12.26783,9561348,9992719}, ellipsiods, zonotopes \citep{10558853,trapiello2023verificationneuralnetworkcontrol,10.1007/978-3-030-53288-8_1}, support functions, and Taylor models \citep{schilling,10.1145/3302504.3311806}, or make linear modeling approximations \citep{10.1007/978-3-031-65112-0_5}.
However, these representations and the methods used to propagate them can incur significant approximation costs, resulting in conservative analyses that can fail to certify safety.
In Appendix \ref{app:baseline}, we empirically quantify this conservatism by comparing RoVer-CoRe against NNV 2.0 \citep{nnv2_cav2023}, a popular verification toolbox for NN control systems (NNCS).
Alternative approaches have been proposed to compute tighter set approximations, but they are restricted to certain classes of system dynamics and controllers.
We propose to overcome these limitations by replacing approximate set propagation tools with HJ reachability analysis, which aims to compute optimal reachable sets.

\subsection{Certificate-Based Methods}
Recent works extend certificate-based methods, particularly control barrier functions (CBFs), to verify safety under state uncertainty.
For example, Measurement-Robust CBFs (MR-CBFs) strengthen the standard CBF inequality using bounds on the estimation error to guarantee safety for all states consistent with that bound \citep{dean2021guaranteeing, cosner2021measurement}.
However, an MR-CBF is not always guaranteed to exist, depending on the magnitude of the error bound.
Robust CBFs (R-CBFs) improve upon MR-CBFs by introducing a robustifying term that handles estimation errors without requiring a fixed error bound a priori \citep{nanayakkara2025safety}.
Most recently, an adaptive extension to R-CBFs has been proposed to reduce conservatism and infeasibility \citep{das2025safe}.
Despite this progress, existing CBF-based methods suffer from common limitations, including key challenges with global feasibility and barrier construction, which can lead to safety violations in practice.
In contrast, our proposed framework provides a globally constructive mechanism that can handle general nonlinear systems and failure set geometries. 

\subsection{HJ Reachability Analysis}\label{sec:related_works_hj}
Our proposed framework builds upon HJ reachability analysis, which computes the optimal reachable set and associated controller under worst-case disturbances that capture model mismatches, external forces, or other adversarial effects \citep{bansal2017hamilton}.
At a first glance, it may appear natural to treat perceptual uncertainty as another such adversarial disturbance.
However, the standard HJ formulation assumes that the controller has perfect state information, optimizing the control input based on the true system state.
Relaxing this assumption fundamentally alters the structure of the underlying differential game, requiring a reformulation of the value function to accomodate partial observability, which is a direction we leave for future work.
In this paper, we instead focus on verifying closed-loop systems induced by fixed perception-based controllers, which, perhaps surprisingly, aligns naturally with the existing HJ reachability framework after appropriate adjustments.
To the best of our knowledge, RoVer-CoRe is the first framework to use HJ reachability to provide formal guarantees for systems operating under perceptual uncertainty.
\section{Problem Setup}\label{sec:problem_setup}

We model a perception-based system with state $x \in \mathcal{X}$, control $u \in \mathcal{U}$, disturbance $d \in \mathcal{D}$, and dynamics $\dot{x} = f(x, u, d)$ governing how $x$ evolves over time until a final time horizon $T$.

The system observation (or output) $o \in \mathcal{O}$ is given by a general observation map $p: o = p(x)$ which is a function of the underlying state $x$.
For example, $o$ can correspond to an image taken by an onboard camera sensor or the output of a motion capture system.
We assume that the system obtains a state estimate $\hat{x} \in \mathcal{X}$ by a general estimator $q: \hat{x} = q(o, e)$ which is a function of the observation $o$ and a perceptual error input $e \in \mathcal{E}$.
For example, $q$ can represent a noisy state-estimation filter or an NN-based state-estimation module that contains learning errors.
Let $\pi: \mathcal{X} \rightarrow \mathcal{U}$ denote a state-based controller which maps the state estimate to a control input, i.e., $u=\pi(\hat{x})$.
See Figure \ref{fig:system} for an illustration of the overall system model.

We denote the set of failure states as $\mathcal{F} \subseteq \mathcal{X}$ (e.g., collision states) which the system is not allowed to enter.
The failure set can be represented by the zero-sublevel set of a Lipschitz-continuous function $l: \mathcal{X} \rightarrow \mathbb{R}$, i.e., $x \in \mathcal{F} \Leftrightarrow l(x) \leq 0$.
Let $\xi^{\pi \circ q(p(\cdot), e(\cdot)), d(\cdot)}_{x_0, t_0}(\tau)$ denote the state achieved at time $\tau \in [t_0, T]$ by starting at initial state $x_0$ at time $t_0$ and applying the control policy $\pi$ over $[t_0, \tau]$ under the error signal $e(\cdot)$ and disturbance signal $d(\cdot)$.
We are interested in verifying the safety of the system starting from state $x_0$ and time $t_0$ under $\pi$ in the presence of the worst-case error signal $e(\cdot)$ and the worst-case disturbance signal $d(\cdot)$ until the finite time horizon $T$.
In other words, we want to verify that $\forall \tau \in [t_0, T], \forall e(\cdot), \forall d(\cdot), \xi^{\pi \circ q(p(\cdot), e(\cdot)), d(\cdot)}_{x_0, t_0}(\tau) \notin \mathcal{F}$.
\section{Background}\label{sec:background}

\subsection{HJ Reachability Analysis}

In this section, we explain HJ reachability analysis in the context of a traditional system verification problem, where there is no perception-based controller.
In Section \ref{sec:approach}, we show how we can extend the framework to verify perception-based controllers.

HJ reachability analysis is concerned with computing the system's Backward Reachable Tube, which we denote as $\text{BRT}$ \citep{lygeros2004reachability, mitchell2005time}.
We define $\text{BRT}$ as the set of all initial states $x \in \mathcal{X}$ starting from which, for all control signals $\mathbf{u}(\cdot)$, there exists a disturbance signal $\mathbf{d}(\cdot)$ such that the system will enter the failure set $\mathcal{F}$ within the time horizon $[t_0, T]$:
\begin{equation}
    \text{BRT} \coloneq \{ x \in \mathcal{X}: \forall \mathbf{u}(\cdot), \exists \mathbf{d}(\cdot),\exists \tau \in [t_0, T], \xi^{\mathbf{u}(\cdot),\mathbf{d}(\cdot)}_{x,t_0}(\tau) \in \mathcal{F} \}.
\end{equation}
The BRT complement precisely captures the set of states for which we can guarantee system safety.

In HJ reachability, computing $\text{BRT}$ is formulated as a robust optimal control problem.
First, we implicitly represent the failure set $\mathcal{F}$ by a failure function $l(x)$ whose zero-sublevel set yields $\mathcal{F}$: $\mathcal{F} = \{ x \in \mathcal{X}: l(x) \leq 0 \}$.
$l(x)$ is commonly the signed distance function to $\mathcal{F}$.
Next, we define the cost function corresponding to a control signal $\mathbf{u}(\cdot)$ and disturbance signal $\mathbf{d}(\cdot)$ to be the minimum of $l(x)$ over the trajectory starting from a state $x$ and time $t$:
\begin{equation}
    J_{\mathbf{u}(\cdot), \mathbf{d}(\cdot)}(x, t) \coloneq \min_{\tau \in [t, T]}l(\xi^{\mathbf{u}(\cdot),\mathbf{d}(\cdot)}_{x,t}(\tau)).
\end{equation}
Since the control aims to avoid $\mathcal{F}$ under worst-case disturbance, the value function corresponding to this robust optimal control problem is:
\begin{equation}
    \label{eq:V}
    V(x, t) \coloneq \max_{\mathbf{u}(\cdot)} \min_{\mathbf{d}(\cdot)} J_{\mathbf{u}(\cdot), \mathbf{d}(\cdot)}(x, t).
\end{equation}
By defining our optimal control problem in this way, we can recover $\text{BRT}$ using the value function.
The value function being non-positive implies that the failure function is non-positive somewhere along the optimal trajectory, or in other words, that the system will inevitably enter $\mathcal{F}$.
Conversely, the value function being positive implies that there exists a control signal that will prevent the system from entering $\mathcal{F}$ even under the worst-case disturbance signal.
Thus, $\text{BRT}$ is computed as the zero-sublevel set of the value function:
\begin{equation}
    \label{eq:BRT}
    \text{BRT} = \{ x \in \mathcal{X}: V(x, t_0) \leq 0 \}.
\end{equation}
Using the principles of optimality and dynamic programming, it can be shown that the value function in Equation \eqref{eq:V} can be computed as the solution to the following final value Hamilton-Jacobi-Isaacs Variational Inequality (HJI-VI):
\begin{align}
    \min\{ D_{t}V(x, t) + H(x, t, \nabla V(x, t)), l(x) - V(x, t) \} = 0, \nonumber \\
    V(x, T) = l(x), \quad \forall t \in [t_0, T].
\end{align}
$D_{t}V(x, t)$ and $\nabla V(x, t)$ represent the temporal derivative and spatial gradient of the value function $V(x, t)$, respectively.
The Hamiltonian $H(x, t, \nabla V(x, t))$ encodes how the control and disturbance interact with the system dynamics:
\begin{equation}\label{eq:hamiltonian}
    H(x, t, \nabla V(x, t)) \coloneq \max_{u \in \mathcal{U}} \min_{d \in \mathcal{D}} \nabla V(x, t) \cdot f(x, u, d).
\end{equation}
The value function in Equation \eqref{eq:V} also induces the optimal safety controller:
\begin{equation}
    \label{eq:u*}
    \mathbf{u}^*(x, t) \coloneq \text{arg}\max_{u \in \mathcal{U}} \min_{d \in \mathcal{D}} \nabla V(x, t) \cdot f(x, u, d).
\end{equation}
Intuitively, the optimal safety controller aligns the system dynamics in the direction of the value function's gradients, thus steering the system towards higher-value states, i.e., away from $\mathcal{F}$.
It can be shown that safety is guaranteed despite worst-case disturbances if the system starts outside of $\text{BRT}$ and applies the control in Equation \eqref{eq:u*} at the $\text{BRT}$ boundary \citep{10665911}.

Unfortunately, the analysis above assumes that the controller has access to the true system state.
When perceptual uncertainty is present, this assumption is violated, so the classical reachability formulation cannot be applied directly.
This motivates our use of a different closed-loop system abstraction induced by a perception-based controller, which we introduce next for the safety analysis.
\section{Approach}\label{sec:approach}

As discussed above, traditional HJ reachability frameworks are not equipped to handle perceptual uncertainty directly.
To address this limitation, we introduce RoVer-CoRe.
Section \ref{sec:abstraction} presents the key closed-loop system abstraction that enables the use of HJ reachability tools, under which the main technical challenge becomes optimizing the closed-loop Hamiltonian.
Section \ref{sec:hamiltonian_bounds_approach} proposes methods to efficiently compute or bound this Hamiltonian and examines the resulting safety guarantees.
Finally, Section \ref{sec:controller_design} discusses how the verification results support robust controller design.

\subsection{Closed-Loop System Abstraction}\label{sec:abstraction}

To begin our analysis, we observe that for a fixed controller, the only mutable inputs to the system are the disturbance signal $d(\cdot)$ and error signal $e(\cdot)$.
Treating both as adversarial, we aim to apply HJ reachability tools to the resulting closed-loop behavior.
Our \textit{key idea} is to concatenate the controller, observation map, and state estimator to form the closed-loop dynamics:
\begin{equation}
    h(x, d, e) \coloneqq f(x, \pi(q(p(x), e)), d),
\end{equation}
whose only external inputs are $d$ and $e$.
This abstraction is illustrated in Figure \ref{fig:system}.

Under this representation, the system interface aligns with traditional HJ reachability methods.
The main difficulty, however, lies in evaluating the closed-loop Hamiltonian, which becomes:
\begin{equation}\label{eq:closed_loop_hamiltonian}
    H(x, t, \nabla V(x, t)) = \min_{d \in \mathcal{D}, e \in \mathcal{E}} \nabla V(x, t) \cdot h(x, d, e).
\end{equation}
Due to the complexity of the controller $\pi(\cdot)$, observation map $p(\cdot)$, and estimator $q(\cdot)$, the Hamiltonian can be a non-convex function of $d$ and $e$, making Equation \eqref{eq:closed_loop_hamiltonian} challenging to solve.
This difficulty is the central obstacle in verifying closed-loop systems under perceptual uncertainty.
In the next section, we propose methods to address this challenge.

\subsection{Computing or Bounding the Closed-Loop Hamiltonian}\label{sec:hamiltonian_bounds_approach}

In this section, we propose methods to efficiently compute or bound the Hamiltonian in Equation \eqref{eq:closed_loop_hamiltonian} and discuss their implications on the computed guarantees.
Users of RoVer-CoRe should determine which of the proposed methods best suits their particular verification problem.

As a first step, we propose to assume that the state estimate $\hat{x}$ is the result of a bounded additive perturbation to the underlying state, i.e., $\hat{x} = q(p(x), e) = x + e$.
This step can be taken without loss of generality, since the noise bounds can always be chosen large enough to contain all perceptual errors encountered during deployment, albeit at the cost of conservatism.
This assumption greatly simplifies the application of the proposed methods described next.
We defer the treatment of more general observation maps and state estimators, such as via generative NNs, to future work.

\subsubsection{Exactly Computing the Hamiltonian}\label{sec:closed_loop_hamiltonian_exact}

In some cases, particularly when the controller is enumerable or has a simple structure, e.g., a linear feedback controller, the optimization problem in Equation \eqref{eq:closed_loop_hamiltonian} can be solved exactly.
In such settings, we can apply HJ reachability tools directly to the closed-loop abstraction with no conservatism.
The verification process also yields the corresponding worst-case disturbance and error signals, which serve as concrete counterexamples when safety is violated.
We illustrate this procedure for the taxiing controller in Section \ref{sec:taxinet} and the grid-based MPC in Section \ref{sec:navigation}.

\subsubsection{Lower Bounding the Hamiltonian}\label{sec:closed_loop_hamiltonian_bound}

When it is not possible to exactly solve the optimization problem in Equation \eqref{eq:closed_loop_hamiltonian} due to the complexity of the controller, we propose to instead compute a lower bound on the closed-loop Hamiltonian, which will result in a conservative yet sound verification of the system \citep{choi2025data}.
Specifically, let $\underline{\mathcal{U}}(x)$ and $\overline{\mathcal{U}}(x)$ represent element-wise lower and upper bounds on the controller output at state $x$ under perceptual uncertainty, i.e., $\underline{\mathcal{U}}(x) \leq \pi\left(q\left(p\left(x\right),e\right)\right) \leq \overline{\mathcal{U}}(x), \forall e\in\mathcal{E}$.
Let $\mathcal{U}(x)$ denote the hyperrectangle defined by $\underline{\mathcal{U}}(x)$ and $\overline{\mathcal{U}}(x)$.
We can lower-bound Equation \eqref{eq:closed_loop_hamiltonian}:
\begin{equation}\label{eq:closed_loop_hamiltonian_with_control_bounds}
    H(x, t, \nabla V(x, t)) \ge \min_{d \in \mathcal{D}, u \in \mathcal{U}(x)} \nabla V(x, t) \cdot f(x, u, d),
\end{equation}
which takes a form similar to the open-loop Hamiltonian in Equation \eqref{eq:hamiltonian} and thus aligns with existing HJ reachability tools.
Intuitively, we robustly handle perceptual uncertainty by capturing all possibilities with $\mathcal{U}(x)$.
$\mathcal{U}(x)$ can be computed efficiently depending on the form of the controller.
For example, if the controller is represented on a grid, we can compute $\mathcal{U}(x)$ by enumeration.
If the controller is an NN, we can compute $\mathcal{U}(x)$ using state-of-the-art NN verification tools.
We illustrate this procedure for the NN-based controller in Section \ref{sec:navigation}.
\begin{remark}
The conservatism of the analysis will scale with the degree with which $\mathcal{U}(x)$ overapproximates the true controller output space.
Although we do not explore it here, we suggest that users can trade off conservatism with computational complexity depending on the characteristics of the system, e.g., by representing $\mathcal{U}(x)$ as a convex hull rather than a hyperrectangle, or by budgeting more computational resources for tighter NN verification.
\end{remark}

\subsection{Robust Controller Design}\label{sec:controller_design}

As described above, RoVer-CoRe evaluates the safety margins of fixed perception-based controllers under perceptual uncertainty.
Here, we discuss how these verification results can be used in practice to improve controller robustness.
In particular, we propose using RoVer-CoRe for safety-guided hyperparameter optimization.
Consider a family of controllers $\pi_\alpha$ parameterized by tunable hyperparameters $\alpha$.
For each $\pi_\alpha$, RoVer-CoRe computes a robust, parameter-conditioned value function $V_\alpha$.
By evaluating $V_\alpha$ over a range of $\alpha$, we can select $\alpha^*$ which maximizes a desired safety objective, such as the safety margin at a given initial state $x_0$, or the volume of the safe set.
This enables safety-guaranteed controller design, which we demonstrate in the next section.
\section{Case Studies}\label{sec:case_studies}

\subsection{Aircraft Taxiing}\label{sec:taxinet}

For our first case study, we analyze a lane-following aircraft taxiing controller \citep{taxinet}.
The aircraft evolves according to:
\begin{align}\label{eq:dynamics}
    \dot{p_x} &= v \sin\theta  &  \dot{p_y} &= v \cos\theta & \dot{\theta} &= u,
\end{align}
where $p_x$ is the crosstrack error, $p_y$ is the downtrack position, and $\theta$ is the heading error relative to the centerline.
$v$ is the linear velocity fixed at $5$ m/s, and $u$ is the commanded angular velocity.
The aircraft starts at $x_0=(0, 100, 0)$ and must follow the centerline as closely as possible using images from a wing-mounted camera.
A deep neural network (DNN) module processes the images to predict the crosstrack error $\hat{p}_x$ and the heading error $\hat{\theta}$.
Using these estimates, the controller computes
$u \coloneq \tan(a\cdot\hat{p}_x + b\cdot\hat{\theta})$, where $a, b \leq 0$ are proportional gains on $\hat{p}_x$ and $\hat{\theta}$, respectively.
The aircraft violates safety if it leaves the runway, corresponding to the failure set $\mathcal{F} = \{x: |p_x| \geq 10\}$.

\begin{figure}[!h]
\centering
\includegraphics[width=0.8\textwidth]{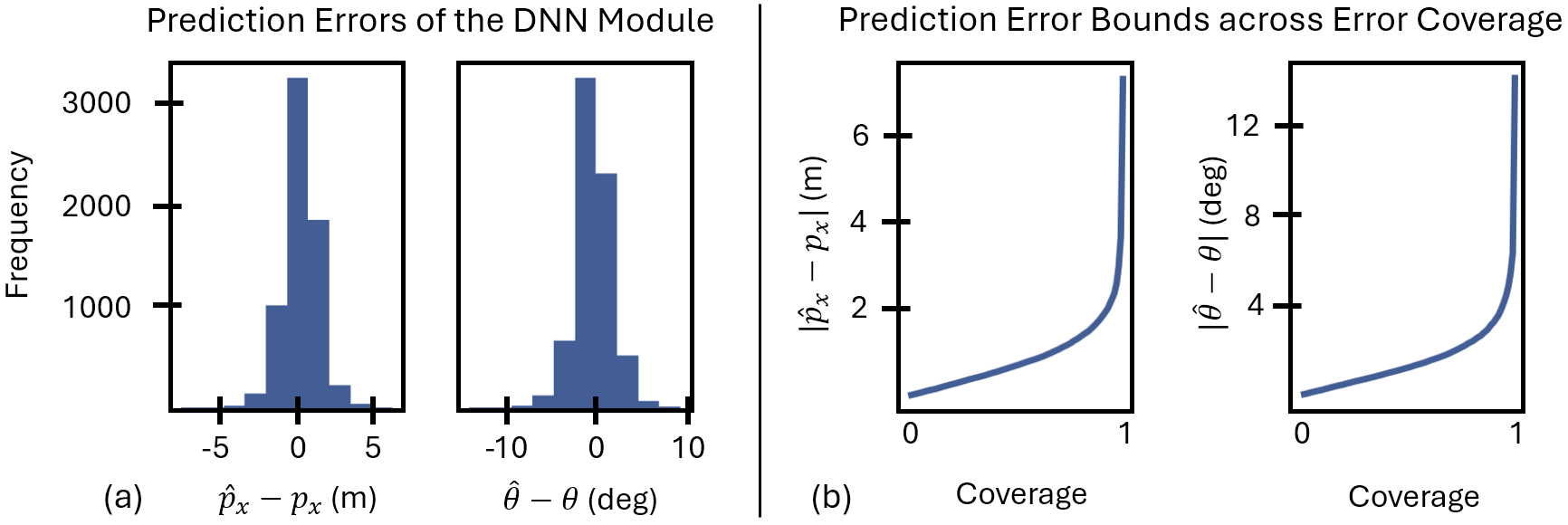}
\caption{(a) Histograms of the prediction errors for $\hat{p}_x$ and $\hat{\theta}$ by the aircraft DNN module. (b) Prediction error bounds for different error coverages.}
\label{fig:errors}
\vspace{-0.2em}
\end{figure}

The DNN predictions can include errors $e_{\hat{p}_x}=\hat{p}_x-p_x,\: e_{\hat{\theta}}=\hat{\theta}-\theta$ that induce system failure.
Thus, we would like to formally verify the system under worst-case perceptual uncertainty.
In Figure \ref{fig:errors}, we plot the error distribution over a test dataset of runway images generated in the X-Plane simulator \citep{xplane}.
Using the error distribution, we compute element-wise norm bounds $\bar{e}_{\hat{p}_x}, \bar{e}_{\hat{\theta}}$ that correspond to different coverages $0 \le c \le 1$ of the observed errors.

\begin{figure}[!h]
\centering
\includegraphics[width=1.0\textwidth]{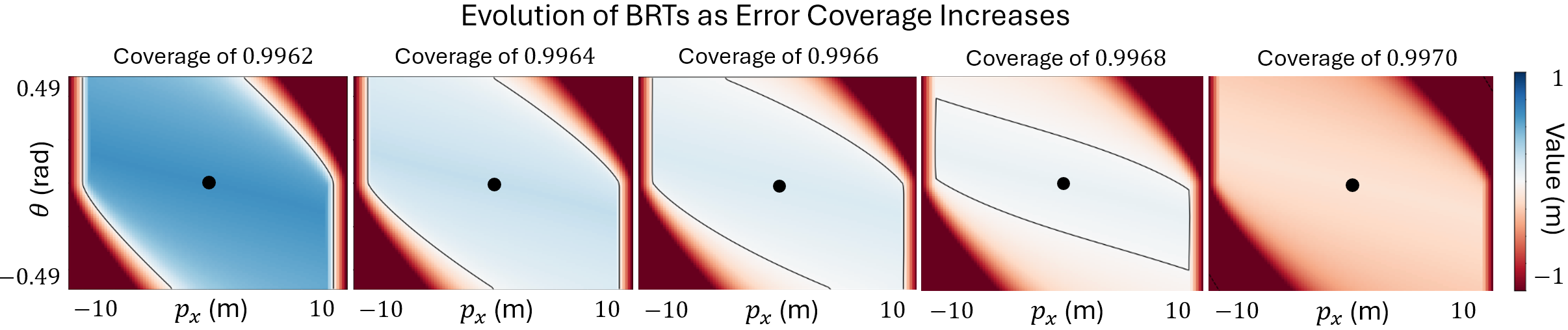}
\caption{Evolution of the BRTs for the taxiing controller as the error coverage increases. Safety is guaranteed in the blue region. (Black) The BRT boundary. (Dot) The initial state.}
\label{fig:taxinet_brts}
\end{figure}

Next, we demonstrate how RoVer-CoRe can be used to verify system safety.
We consider the controller $a = -0.013, b = -0.44$, as suggested in \citet{10075409}.
Since the controller is sufficiently simple, we can compute the closed-loop Hamiltonian and conduct an exact verification: $H(x, t, \nabla V(x, t))=\beta_1v\sin\theta+\beta_2v\cos\theta+\beta_3\tan\left( a\left( p_x+e^*_{\hat{p}_x} \right) + b\left( \theta+e^*_{\hat{\theta}} \right) \right)$, where $\nabla V(x,t)=[\beta_1, \beta_2, \beta_3]^T$, and the worst-case errors are given by $e^*_{\hat{p}_x}=-\sgn(\beta_3a)\bar{e}_{\hat{p}_x},\:e^*_{\hat{\theta}}=-\sgn(\beta_3b)\bar{e}_{\hat{\theta}}$.
We compute the robust value functions across different coverages of the observed test errors using the grid-based \texttt{hj\_reachability} Python toolbox \citep{hj_reachability_python} over the state space $[-11, 11]$ m $\times$ $[100, 250]$ m $\times$ $[-0.49, 0.49]$ rad with a grid shape of $[101, 101, 101]$ for a time horizon of $20$ s.
Each value function takes $\leq 2$ s to compute on an NVIDIA 3090 Ti GPU.
In Figure \ref{fig:taxinet_brts}, we plot the evolution of the BRTs as the error coverage increases.
The plots show that we can formally verify the safety of the initial state of the system up to an error coverage of $0.9968$.

\begin{figure}[!h]
\centering
\includegraphics[width=0.8\textwidth]{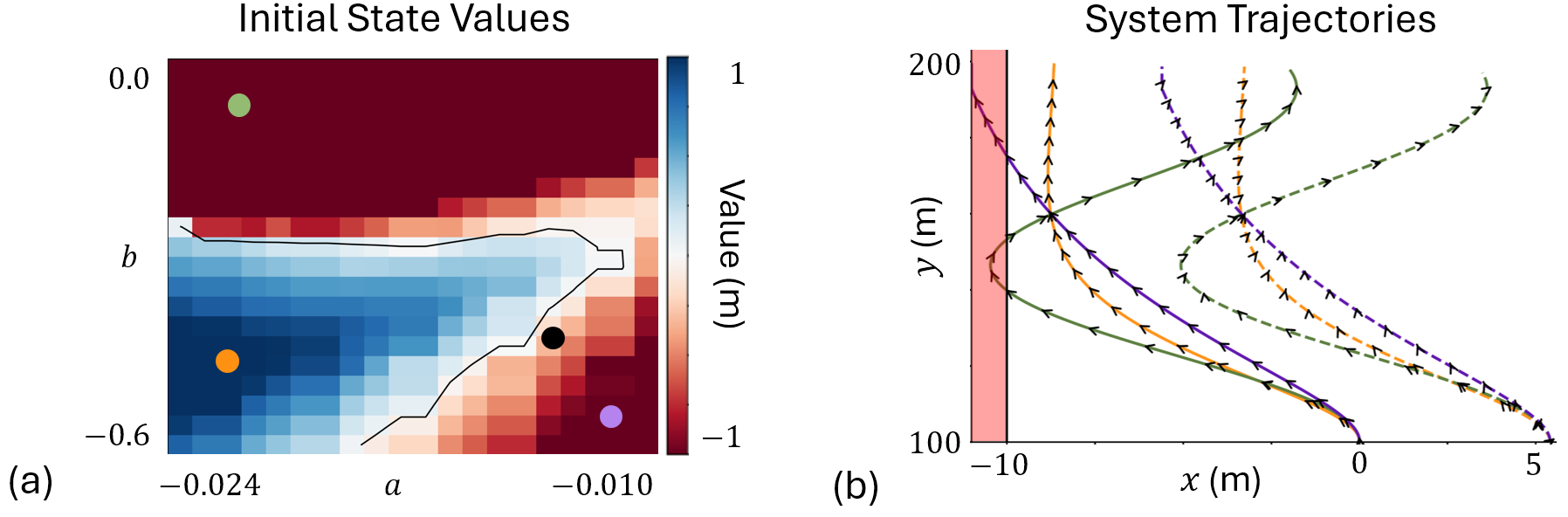}
\caption{(a) Value function at the initial state under an error coverage of $0.997$ across the controller space.
(b) Actual (solid) and perceived (dashed) trajectories under worst-case uncertainty.}
\label{fig:design}
\end{figure}

Finally, we demonstrate how RoVer-CoRe can guide robust controller design.
Figure \ref{fig:design} (a) shows the safety value of the initial state under different controllers across the 2D controller-hyperparameter space for an error coverage of $0.997$.
The original controller (black dot) is unsafe, whereas controllers such as the orange point within the blue region remain robust to the desired level of perceptual uncertainty.
Figure \ref{fig:design} (b) visualizes the corresponding actual (solid) and perceived (dashed) trajectories: the orange controller preserves safety under worst-case perception errors.

To illustrate how RoVer-CoRe exposes failure modes, we also analyze the unsafe controllers marked in green and purple.
The green controller's high cross-track gain causes over-correction when the aircraft misperceives its lateral position, driving it outside the runway limits.
The purple controller's large heading-error gain leads the aircraft to believe it is aligned with the centerline while actually drifting off course.
These results highlight how RoVer-CoRe enables both formal safety guarantees and diagnostic insights for principled controller refinement.

\subsection{NN-Based Rover Navigation}\label{sec:navigation}

Our second case study considers a rover navigating around obstacles under perceptual uncertainty, inspired by NASA's \textit{Endurance} concept for long-range lunar night exploration \citep{baker2024endurance}.
The rover uses visual odometry whose error grows in low-light conditions.
Uncertainty can be reset to zero by turning on headlights, but this comes at a high energy cost.
We model the error with an element-wise norm bound that grows linearly with the time $t'$ since the lights were last activated: $\bar{e}^{(t')}=(\bar{e}_{\hat{p}_x}^{(t')}, \bar{e}_{\hat{p}_y}^{(t')}, \bar{e}_{\hat{\theta}}^{(t')})=(0.1t', 0.1t', 0.02t')$.
The rover follows Equation \eqref{eq:dynamics} with $v=1$ m/s and violates safety upon collision.
Since safety depends on the closed-loop evolution of the time-varying perceptual uncertainty, the problem is a natural fit for RoVer-CoRe.

\begin{figure}[!h]
\centering
\includegraphics[width=1.0\textwidth]{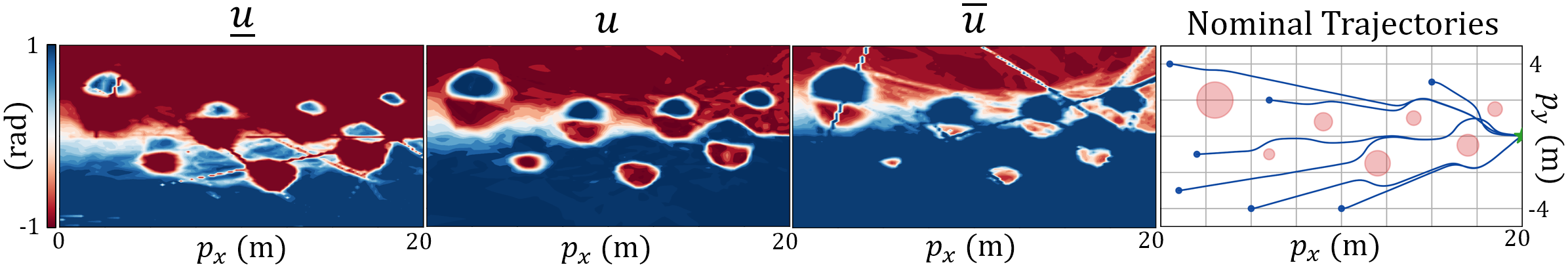}
\caption{From left to right: the lower bounds, nominal outputs, and upper bounds of the NN at ($\theta=0$ rad, $t=1.5$ s). Rightmost: Rollouts by the NN-based controller successfully reach a goal (green star) while avoiding obstacles (red circles) under perfect state estimation.}
\label{fig:controller}
\end{figure}

Next, we use RoVer-CoRe to verify the rover's safety following Section \ref{sec:closed_loop_hamiltonian_bound}.
To show the applicability of our method, we evaluate two controllers: an expert MPC defined on a grid, for which control bounds are obtained directly via enumeration, and an NN trained to imitate the MPC, for which bounds are computed using the $\alpha,\beta$-CROWN verifier \citep{10.5555/3540261.3542550}.
We compute bounds assuming the lights are off on a state-time grid of $[0, 20]$ m $\times$ $[-5, 5]$ m $\times$ $[-\pi, \pi]$ rad $\times$ $[0, 5]$ s with shape $4 \times [100]$, which takes $\approx30$ minutes on an NVIDIA 5090 GPU.
The NN bounds at ($\theta=0$ rad, $t=1.5$ s), along with rollouts under perfect state estimation, are shown in Figure \ref{fig:controller}.

\begin{figure}[!h]
\centering
\includegraphics[width=1.0\textwidth]{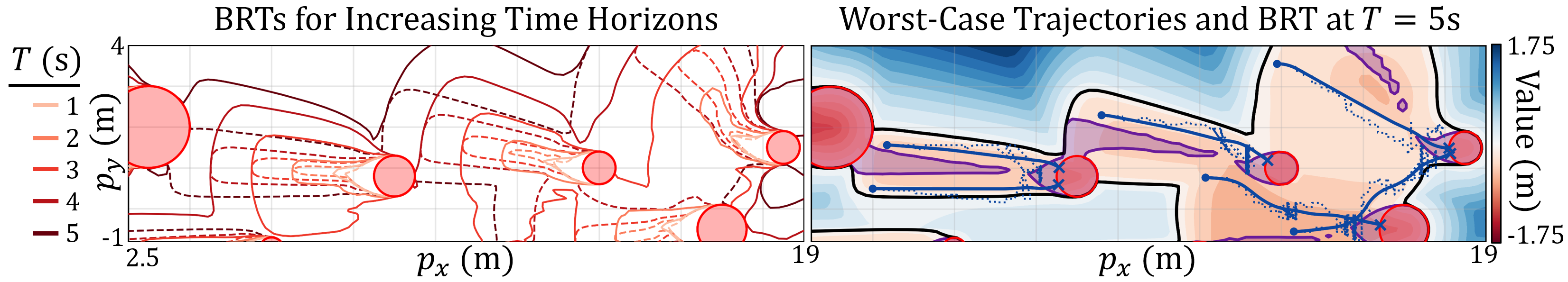}
\caption{Left: BRT boundaries computed by RoVer-CoRe when the lights are off for the original MPC (dashed) and the NN-based controller (solid) as the time horizon $T$ increases (color). Right: The initial-time value function for $T=5$s for the original MPC as computed by RoVer-CoRe (color) compared with the Monte Carlo baseline (purple). True (solid) and perceived (dotted) worst-case rollouts that lead to collision are shown in blue.}
\label{fig:brts}
\end{figure}

After obtaining the controller bounds, we compute the corresponding robust value function using the grid-based \texttt{hj\_reachability} Python toolbox \citep{hj_reachability_python} on the same grid as above, which takes $\approx20$ s on an NVIDIA 5090 GPU.
The BRTs for increasing time horizons $T$ for the NN (solid) and MPC (dashed) appear in the left plot of Figure \ref{fig:brts}.
The NN BRTs are consistently more inflated than those of the MPC, reflecting both imitation error and the looseness of the $\alpha,\beta$-CROWN bounds.
This yields a conservative but sound guarantee; starting from any state outside these BRTs is guaranteed to be safe under worst-case perceptual uncertainty.

To evaluate efficiency and completeness, we compare RoVer-CoRe against a naive Monte Carlo baseline for verifying the MPC.
For each grid state, we roll out $10^4$ trajectories under uniformly sampled state uncertainties and record the minimum safety value observed.
The estimates stabilize within collecting $\frac{1}{10}$ of the samples.
The resulting BRT estimate (purple, right plot of Figure \ref{fig:brts}) misses many unsafe states that RoVer-CoRe correctly identifies.
Since we compute the Hamiltonian for the MPC exactly via enumeration, RoVer-CoRe also produces counterexamples, several of which are shown in the same plot.
These results underscore the rigor and completeness of our method.
In Appendix \ref{app:baseline}, we further compare against the popular set-based NNV 2.0 tool \citep{nnv2_cav2023}.

\begin{figure}[!h]
\centering
\includegraphics[width=1.0\textwidth]{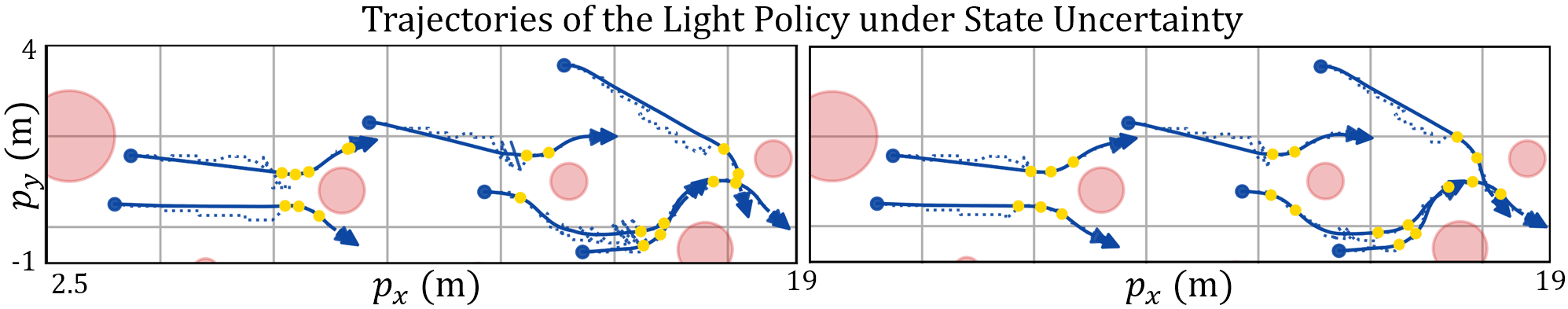}
\caption{True (solid) and perceived (dotted) worst-case rollouts starting from the same initial states as in Figure \ref{fig:brts}, under the original MPC (left) and NN-based controller (right), which would be unsafe without lights. The light policies derived by RoVer-CoRe activate the lights (yellow dots) only when needed to guarantee safety under worst-case uncertainty.}
\label{fig:policy}
\end{figure}

Finally, we show how RoVer-CoRe enables robust controller design.
If the rover kept its lights on continuously, safety would be governed by the zero-uncertainty BRT.
Leveraging this, we treat the zero-uncertainty BRT as a surrogate failure set and, for each state, use RoVer-CoRe to compute the earliest time at which the growing-uncertainty dynamics would drive the rover into this set.
This time corresponds exactly to the maximum duration the rover can safely operate with the lights off.
Using these durations, we construct a light-activation policy that guarantees safety while reducing unnecessary energy use.
Figure \ref{fig:policy} shows the resulting policy applied to the same initial states that previously led to failure: the rover remains safe, and the lights activate only when needed.
As expected, the policy triggers the lights slightly earlier for the NN-based controller, reflecting its weaker robustness.
For both controllers, the rover is able to navigate in the dark for significant periods while retaining formal safety guarantees.
These results highlight how RoVer-CoRe enables both rigorous safety verification and practical controller design in challenging uncertainty settings.

\section{Conclusion}\label{sec:conclusion}

We introduced RoVer-CoRe, the first HJ reachability-based framework to verify perception-based controllers operating under bounded perceptual uncertainty by formulating a closed-loop abstraction that unifies perception, estimation, and control.
A central insight is the characterization of the closed-loop Hamiltonian, for which we develop exact and conservative bounding methods that enable safety verification for different forms of controllers.
Through case studies, with code open-sourced\textsuperscript{\ref{note:code}}, we show that RoVer-CoRe not only provides formal safety guarantees but also produces significantly tighter guarantees than set-based NNCS verification tools (Appendix \ref{app:baseline}), exposes perception-induced failure modes, and supports robust controller design.
Future directions include incorporating probabilistic uncertainty models, improving scalability via learning-based reachability tools, and extending the framework to richer perceptual pipelines and controller synthesis.


\acks{This work was supported in part by a NASA Space Technology Graduate Research Opportunity, the NSF CAREER Program under award 2240163, and the DARPA ANSR program. Part of the research was carried out at the Jet Propulsion Laboratory, California Institute of Technology, under a contract with the National Aeronautics and Space Administration (80NM0018D0004).}

\bibliography{bibliography/bansal_papers, bibliography/reachability, bibliography/references}

\appendix
\section{Comparison with Set-Based NNCS Verification}\label{app:baseline}

To empirically validate the conservatism reduction discussed in Section \ref{sec:related_works}, we compare RoVer-CoRe against NNV 2.0 \citep{nnv2_cav2023,10.1007/978-3-030-53288-8_1}, a popular set-based neural network control system (NNCS) verification tool.
We design a simplified experiment to enable a fair comparison between the two methods.

\paragraph{Setup.}
Both methods verify a rover following Equation \eqref{eq:dynamics} with $v=1$ m/s and control $u \in [-1, 1]$ rad/s over the state space $[0, 20]$ m $\times$ $[-5, 5]$ m $\times$ $[-\pi, \pi]$ rad with a grid shape of $[100]^3$.
A single circular obstacle of radius $1$ m is placed at $(p_x, p_y) = (16, 0)$ on the rover's nominal path toward the goal at $(20, 0)$.
A pure ReLU MLP controller ($3 \to 128 \to 128 \to 1$) is trained via supervised imitation of an MPC and shared identically by both methods.
Perception uncertainty is bounded by $\bar{e} = (\bar{e}_{\hat{p}_x}, \bar{e}_{\hat{p}_y}, \bar{e}_{\hat{\theta}}) = (0.5\text{ m}, 0.5\text{ m}, 0.1\text{ rad})$; the NN observes the perturbed state $\hat{x} = x + e$ with $|e| \leq \bar{e}$, while the dynamics evolve under the true state $x$.

\paragraph{NNV method.}
NNV computes the BRT via forward reachability under time-reversed dynamics, which is mathematically equivalent to backward reachability in continuous time under the single-player (fixed controller, worst-case disturbance) setting.
The NN controller is analyzed using NNV's approx-star method, which propagates Star sets through the ReLU layers to bound the control output \citep{10.1007/978-3-030-53288-8_1}.
The nonlinear plant dynamics are propagated using CORA's zonotope-based reachability.
Following NNV's documented pipeline, a per-step interval hull is applied to maintain a single set representation at each time step.
The initial obstacle set is partitioned into $100$ slices along $\theta$ to keep trigonometric intervals tight.

\begin{figure}[h]
\centering
\includegraphics[width=0.35\textwidth]{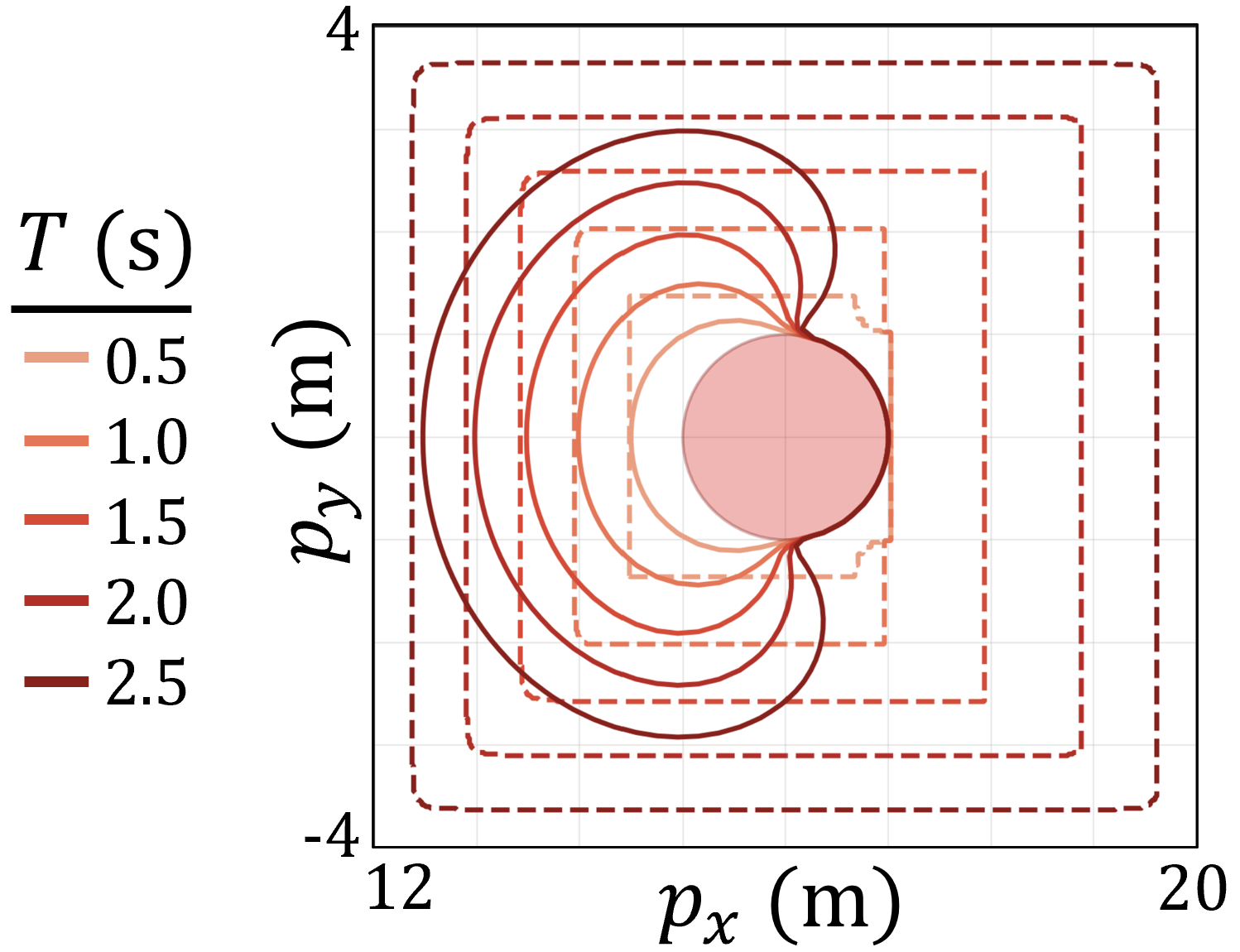}
\vspace{-1em}
\caption{BRT boundaries at $\theta=0$ for increasing time horizons $T$ (color). RoVer-CoRe (solid) produces smooth, tight boundaries, while NNV v2.0 (dashed) produces compositionally rectangular, bloated boundaries due to the wrapping effect. Circle: obstacle.}
\label{fig:baseline}
\vspace{-1em}
\end{figure}

\paragraph{Results.}
Table \ref{tab:baseline} compares BRT volumes across time horizons.
NNV produces BRTs that are $1.9$--$3.2\times$ more conservative than RoVer-CoRe's, with the ratio growing as the overapproximation errors from interval-hull set propagation accumulate through the nonlinear dynamics, commonly known as the wrapping effect.
Figure \ref{fig:baseline} visualizes the BRT boundaries at $\theta=0$ for both methods.
RoVer-CoRe's BRTs exhibit smooth boundaries that closely follow the nonlinear dynamics, whereas NNV's BRTs are compositionally rectangular and visibly bloated.
The full NNV computation (all time horizons) takes $\approx 15$ minutes on $24$ CPU cores; the RoVer-CoRe pipeline takes $\approx 17$ s for $\alpha,\beta$-CROWN bounds \citep{10.5555/3540261.3542550} and $\approx 10$ s for the HJ PDE solve \citep{hj_reachability_python} on a single GPU.

\begin{table}[h]
\centering
\small
\caption{BRT volume comparison. NNV uses $100$ $\theta$-partitions with per-step interval hull.}
\label{tab:baseline}
\vspace{0.5em}
\begin{tabular}{r|rr|r}
\hline
$T$ (s) & RoVer-CoRe (m$^2{\cdot}$rad) & NNV v2.0 (m$^2{\cdot}$rad) & Ratio \\
\hline
0.5 & 27.4 & 50.9 & 1.86$\times$ \\
1.0 & 39.9 & 94.9 & 2.38$\times$ \\
1.5 & 58.1 & 162.2 & 2.79$\times$ \\
2.0 & 82.8 & 260.9 & 3.15$\times$ \\
2.5 & 115.3 & 365.3 & 3.17$\times$ \\
\hline
\end{tabular}
\end{table}

\paragraph{Discussion.}
The conservatism gap reflects a fundamental difference between set-based propagation and grid-based HJ reachability.
Set-based methods must represent the reachable set at each time step using a finite geometric object (e.g., a box or zonotope), and each such overapproximation introduces error that compounds through subsequent steps of the nonlinear dynamics, commonly known as the wrapping effect.
This explains the growing ratio in Table \ref{tab:baseline}: a small per-step overapproximation accumulates into a large conservatism gap over many steps.
HJ reachability, by contrast, solves the value function PDE on a fixed grid without maintaining an explicit set representation, and thus avoids this compounding error entirely.
We also tested a tighter NNV configuration that avoids the per-step bounding box by propagating each set element independently; even in this best case, the BRT volume ratio remains $2.04\times$ at $T=1$ s, though this mode is intractable beyond $T \approx 1$ s due to exponential growth in the number of set elements \citep{10.1007/978-3-030-53288-8_1}.
Full experimental details and reproduction scripts are available in the open-source repository\textsuperscript{\ref{note:code}}.

\end{document}